\documentclass[twoside]{article}

\usepackage[accepted]{aistats2024}

\usepackage[round]{natbib}
\usepackage{bm}
\usepackage{amssymb, amsthm}

\usepackage{graphicx}

\usepackage{physics}
\usepackage{algorithm}
\usepackage{algpseudocode}
\usepackage{apptools}

\newtheorem{define}{Definition}

\newtheorem{assume}[define]{Assumption}
\newtheorem{theorem}[define]{Theorem}

\def\bo{\bm{\Omega}}
\def\bt{\bm{\Theta}}
\def\ec{\mathcal{E}}
\def\ic{\mathcal{I}}
\def\id{\text{id}}
\def\pa{\text{pa}}
\def\tt{\tilde{\bm{\Theta}}}
\def\uc{\mathcal{U}}

\def\e{\bm{e}}
\def\r{\bm{r}}
\def\u{\bm{u}}
\def\x{\bm{x}}

\def\B{\bm{B}}
\def\I{\bm{I}}
\def\J{\bm{J}}
\def\M{\bm{M}}
\def\R{\mathbb{R}}
\def\U{\bm{U}}

\begin{document}

%

%

\twocolumn[

\aistatstitle{Learning Cyclic Causal Models from Incomplete Data}

\aistatsauthor{ Muralikrishnna G. Sethuraman \And Faramarz Fekri}

\aistatsaddress{Georgia Institute of Technology \And Georgia Institute of Technology} ]

\begin{abstract}
Causal learning is a fundamental problem in statistics and science, offering insights into predicting the effects of unseen treatments on a system. Despite recent advances in this topic, most existing causal discovery algorithms operate under two key assumptions: (i) the underlying graph is acyclic, and (ii) the available data is complete. These assumptions can be problematic as many real-world systems contain feedback loops (e.g., biological systems), and practical scenarios frequently involve missing data. In this work, we propose a novel framework, named MissNODAGS, for learning cyclic causal graphs from partially missing data. Under the additive noise model, MissNODAGS learns the causal graph by alternating between imputing the missing data and maximizing the expected log-likelihood of the visible part of the data in each training step, following the principles of the expectation-maximization (EM) framework. Through synthetic experiments and real-world single-cell perturbation data, we demonstrate improved performance when compared to using state-of-the-art imputation techniques followed by causal learning on partially missing interventional data. 
\end{abstract}

\section{INTRODUCTION}
\label{sec:intro}


Understanding causal relations between variables is a fundamental challenge that spans various scientific disciplines \citep{sachs_causal_2005, zhang_integrated_2013, segal_learning_2005, imbens2015causal}. Establishing a causal model allows us to predict how a system will respond to novel disturbances \citep{solus2017consistency, pmlr-v206-sethuraman23a}, while also facilitating precise decision-making \citep{sulik2017encoding}. Typically, causal relations are modeled using \emph{directed graphs} (DG), where the nodes correspond to the variables of interest and the edges encode the causal relationships.     


The existing causal discovery methods typically focus on the case where the available data is complete. This, in practice, often diverges from reality due to the frequent occurrence of missing data in various real-world contexts \citep{little2019statistical}. The mechanism underlying missing data can be broadly categorized into three types: (i) missing at random (MAR), (ii) missing completely at random (MCAR), and (iii) missing not at random (MNAR). In the MNAR scenario, the missingness is contingent on both the observed and the missing samples, while in MAR, the missingness relies solely on the value of the fully observed data. MCAR represents a particular instance of MAR where the cause of missingness is completely random. It is worth noting that certain aspects of MNAR have received attention in the context of causal discovery \citep{gain2018structure, tu2019causal}, while MCAR has been relatively unexplored. 

Some of the previous works deal with missing data by deleting the samples containing missing values \citep{carter2006solutions, 10.5555/3020847.3020865}. While this might work when the missing probability is low, the drop in performance is significantly high as the missing probability increases. Another feasible approach is to first impute the missing data followed by running a causal learning algorithm on the imputed data. However, this may introduce bias in modeling the underlying data distribution \citep{kyono2021miracle}. 


More recently, \citet{gao2022missdag} showed that under the assumption that the underlying graph is a \emph{directed acyclic graph} (DAG), alternating between data imputation and causal learning can lead to significant improvements in learning the graph structure compared to the previous approach. DAG assumption is popular in the causal learning literature as it allows for factorizing the joint distribution into simpler conditional densities, in addition to narrowing down the space of candidate graphs. While this is appealing from a computational standpoint, there is compelling evidence that feedback loops (cycles) exist in many real-world systems \citep{sachs_causal_2005, freimer_systematic_2022}. Such systems generally necessitate interventional data to learn the causal graph, which can be challenging to acquire. Fortunately, recent technological advances, e.g., CRISPR-Cas9 and single-cell RNA-sequencing \citep{dixit_perturb-seq_2016} in biology, have made large-scale interventional data more accessible. Encouraging more causal discovery algorithms to relax reliance on the acyclicity assumption \citep{llc, pmlr-v124-huetter20a, pmlr-v206-sethuraman23a}.

\textbf{Contributions}. In this work, we propose a novel framework called MissNODAGS that is capable of handling cycles in the underlying causal graph in the presence of missing data that is MCAR. Following the footsteps of \citet{gao2022missdag}, MissNODAGS alternates between imputing the missing values, and maximizing the expected log-likelihood of the visible part of the data in each training iteration. Similar to \citet{pmlr-v206-sethuraman23a}, we employ residual normalizing flows \citep{iresnet} for modeling the data likelihood. Through experiments on synthetic and a real-world gene perturbation dataset, we show that our proposed framework achieved improved performance when compared to using state-of-the-art imputation techniques followed by causal learning on MCAR interventional data. 

\textbf{Organization}. In Section \ref{sec:rel-works}, we provide a brief review of the relevant literature on causal discovery from both complete and incomplete data, followed by a description of the problem setup, relevant preliminaries, and the modeling assumptions in Section \ref{sec:prob-setup}. In Section \ref{sec:missnodags}, we present the proposed expectation-maximization based MissNODAGS framework. Finally, MissNODAGS is validated on various synthetic data sets and on a real-world gene perturbation data set in Section \ref{sec:experiments}.

\section{RELATED WORK}
\label{sec:rel-works}

\paragraph{Causal Discovery from Complete Data.} Causal discovery algorithms can be broadly categorized into three groups: \emph{constraint-based methods}, a canonical example here is the PC algorithm \citep{sprites,triantafillou2015constraint,heinze2018invariant}, the main idea here is to estimate the causal graph based on certain constraints enforced by the conditional independencies observed in the data. These methods suffer from poor scalability as it typically require iterating over all possible conditional independences over each pair of variables in the system. Moreover, these algorithms only identify the graph structure up to its \emph{Markov equivalence class}. While the majority of the constraint-based methods rely on the acyclicity assumption on the causal graph structure, there have been a few attempts to extend this framework to allow for cycles \citep{richardson1996discovery}.

The second category, \emph{score-based methods} such as GES \citep{Meek1997GraphicalMS,hauser2012characterization} rely on maximizing a score function over the space of candidate graphs, such that the score is maximized by the true causal graph. Popular choices for score functions include the data likelihood function, posterior likelihood defined with respect to some prior, and the regularized version such as the Bayesian information criteria (BIC). Typically, these methods apply greedy search approaches to circumvent the super-exponential search space. Some notable works that include score based approach for learning cyclic graphs include \citet{llc, pmlr-v124-huetter20a, amendola2020structure, cyclic_equil, 10.1214/17-AOS1602}

In the recent past, \cite{notears} introduced the NOTEARS algorithm that posed DAG learning in the form of a continuous optimization problem when the underlying system is linear. This spurred several extensions \citep{yu2019dag, ng2020role, ng2022masked, zheng20learning, lee2019scaling} for learning DAGs under various assumptions on the underlying causal system in the observational setting. \citet{dcdi} extended this framework to allow for learning DAGs from interventional data. These methods typically rely on using an augmented lagrangian-based solver for the underlying optimization problem, which can be tricky to optimize. \citet{pmlr-v206-sethuraman23a} then adopted this framework for learning nonlinear cyclic directed graphs from interventional data, while also eliminating the need for augmented lagrangian-based solvers by directly modeling the data likelihood.

The final category, \emph{hybrid methods}, combines aspects of both the constraint-based as well as the score-based methods \citep{tsamardinos2006max,solus2017consistency,wang2017permutation}. These methods typically contain a score function that relies on some conditional independence constraints. 

\paragraph{Causal Discovery from Incomplete Data.} 

Causal discovery algorithms that learn from incomplete data typically fall under one of two categories: the first category includes algorithms that only use those samples that are fully observed or ignore the variables containing missing values involved in a conditional independency test \citep{gain2018structure, strobl2018fast, tu2019causal}. However, these methods only work well when the missingness mechanism can be ignored and the number of missing samples is low. In the second category, the missing data is first imputed followed by causal discovery \cite{singh1997learning}. Imputation strategies play a crucial role in these approaches and some notable imputation algorithms include Multivariate Imputation by Chained Equations (MICE) \citep{white2011multiple}, MissForest (MF) \citep{stekhoven2012missforest}, optimal transport (OT)-imputation \citep{muzellec2020missing}, and a few deep learning based approaches \citep{li2018learning, luo2018multivariate}.

More recently, alternating imputation and graph learning strategies have risen in popularity. \citet{wang2020causal} adopted the generative adversarial network based imputation \citep{li2018learning} for learning DAGs by imputing the missing data using a GAN at each training iteration followed by learning the skeleton of the graph. They showed that combining these two modules results in improved performance. \citet{gao2022missdag} devised an expectation maximization-based framework that is capable of learning DAGs from observational data. By utilizing the benefits of a simplified likelihood function arising from the DAG assumption they propose a flexible framework that can be integrated with several causal discovery algorithms. In this work, we extend the alternating imputation and causal learning strategy for learning cyclic causal graphs from incomplete interventional data for both linear and nonlinear structural equation models.

\section{PROBLEM SETUP}
\label{sec:prob-setup}

We now provide a formal description of our problem starting with the notational conventions. All vectors are denoted by lowercase boldface letters, $\x$, matrices are denoted by uppercase boldface letters, $\bm{B}$. Given a set $\mathcal{I}$, $\x_\ic \in \mathbb{R}^{|\ic|}$ denotes a vector containing the components of $\x$ indexed by $\ic$. Similarly, $\bm{B}_{\ic, \uc}$ denotes all the rows of the matrix indexed by $\ic$ and columns indexed by the set $\uc$.

\subsection{Modeling Causal Graphs via Structural Equations}
\label{ssec:sem}

Let $G = (V, E)$ denote a (cyclic) causal graph, where $V = (v_1, \ldots, v_d)$ is the set of vertices, and $E \subseteq V \times V$ denotes the edge set. We associate a random vector $\x = (x_1, \dots, x_d)$ to the nodes of the graph, that is, the random variable $x_i$ corresponds to the node $v_i$. Following the framework proposed by \citet{sem1} and \citet{sem2} the functional relationships between the nodes in $G$ can represented using the \emph{structural equation model} (SEM). That is, 
\begin{equation}
    x_i = f_i\big(\x_{\pa(i)}\big) + e_i, \quad i = 1, \ldots, d,
    \label{eq:sem-indiv}
\end{equation}
where $\pa(i)$ denotes the parent set of $x_i$, $j \in \pa(i)$ if and only if $v_j \to v_i \in E$. The variables $\e = (e_1, \ldots, e_d)$ in \eqref{eq:sem-indiv} are known as the intrinsic noise variables and are assumed to be independent of $\x$. Additionally, we assume that the system is free of unmeasured latent confounders. This implies that the noise variables are independent of each other. Let $f = (f_1, \ldots, f_d)$, the structural equations in \eqref{eq:sem-indiv} can be combined to obtain the following joint form
\begin{equation}
    \x = f(\x) + \e.
    \label{eq:sem-comb}
\end{equation}
Let $\id$ denote the identity map, then $\e = (\id - f)(\x)$. We assume that $(\id - f)^{-1}$ exists, therefore $\x = (\id - f)^{-1}(\e)$. Note that this always holds when the underlying graph is acyclic. 

Interventions can be readily incorporated into \eqref{eq:sem-comb}. In this work, we assume that all interventions are surgical \citep{sprites, sem2}, i.e., all the incoming edges to the intervened nodes are removed. Following the notational convention of \citet{pmlr-v206-sethuraman23a}, let $\mathcal{E} = (\ic, \uc)$ denote an interventional experiment, where $\ic$ denotes the set of intervened nodes and $\uc$ denotes the set of purely observed nodes. Under this setting, the modified SEM now becomes,
\begin{equation}
    \x = \U f(\x) + \U\e + \bm{c},
    \label{eq:sem-inter}
\end{equation}
where $\bm{c} = [\bm{c}_\mathcal{I}, \bm{c}_\mathcal{U}]$ and $\bm{c}_\ic = \x_\ic$ and $\bm{c}_\uc = \bm{0}$, and $\U$ is diagonal matrix such that $U_{ii} = 1$ if and only $v_i \in \uc$, otherwise $U_{ii} = 0$. Let $p_E(\e)$ denote the probability density defined over the vector $\e$, and $\tilde{p}_X(x_\ic)$ denote a probability density function defined over $x_\ic$. Assuming that $(\id - f)$ is differentiable, we then have 
\begin{multline}
    p_X(\x) = \tilde{p}_X(\x_\ic)p_E\Big([(\id - \U f)(\x)]_\uc\Big)\\
        \Big|\det \J_{(\id - \U f)}(\x)\Big|,
    \label{eq:px-interv}
\end{multline}
where $\J_{(\id - \U f)}(\x)$ denotes the Jacobian of $(\id - \U f)$ at $\x$. 

\subsection{Modeling Missing Data}

Our focus is on the setting where a part of the data is missing. Let $\mathcal{D} = \{\x^{(i)}, \r^{(i)}, \uc^{(i)}\}_{i=1}^M$ denote the given data set, where $\x^{(i)} \in \R^d$ is the node observations corresponding to $i$-th sample, $\r^{(i)} \in \R^d$ is a binary vector indicating whether a node has a missing observation in the $i$-th sample, i.e., $r_j^{(i)} = 0$ if $x_j^{(i)}$ is missing and $r_j^{(i)} = 1$ otherwise. Finally, $\uc^{(i)}$ denotes the set of nodes that are not intervened on in the $i$-th sample. We assume that none of the missing nodes in a sample are intervened on. With the MCAR assumption on the missingness mechanism, we have
\begin{equation*}
    P(\x, \r \mid \bm{\theta}, \bm{\gamma}) = P(\x \mid \bm{\theta}) P(\r\mid\bm{\gamma}),
\end{equation*}
where $\bm{\theta}$ denotes the parameters of the density defined over $\x$ and $\bm{\gamma}$ denotes the parameters of the density defined over $\r$. The parameter $\bm{\theta}$ combines the parameters of the function $f$, the noise density $p_E$, and the graph structure $G$. The above equation conveys the fact that the probability of a node being missed is independent of the node variables. This allows us to only focus on $P(\x\mid\bm{\theta})$ in order to learn the causal graph structure. 

Let $\Gamma_i = \{j: r_j^{(i)} = 1\}$ denote the set of indices of the nodes that are not missing in the $i$-th sample, and $\Omega_i = \{1, \ldots, d\}\setminus \Gamma_i$ denote the set of indices of nodes that are missing in the $i$-th sample. Then, 
\begin{equation}
    p_X\Big(\x_{\Gamma_i}^{(i)}\mid \bm{\theta}\Big) = \int p_X\Big(\x_{\Gamma_i}^{(i)}, \x_{\Omega_i}^{(i)} \mid \bm{\theta}\Big) \, d\x_\Omega^{(i)}.
    \label{eq:obs-prob}
\end{equation}
The above integral is in general intractable as it requires marginalizing over the nodes that are missing and often does not have a closed-form solution. In section \ref{sec:missnodags}, we discuss how we overcome the intractability of \eqref{eq:obs-prob} in order to maximize likelihood of the observed nodes. 

\section{MISSNODAGS}
\label{sec:missnodags}

In this section, we present an overview of the MissNODAGS framework, followed by the details on computing the log-likelihood of the combined missing and observed nodes. We then discuss the specifics of imputing the missing nodes for the cases when the SEM is linear and nonlinear.  

\begin{figure*}[t]
    \centering
    \includegraphics[width=0.7\linewidth]{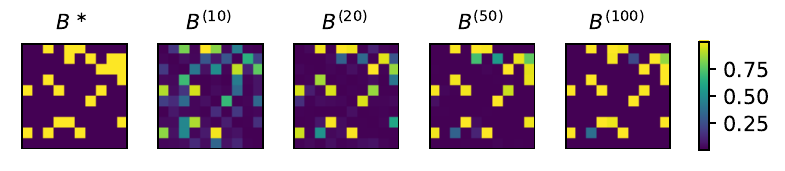}
    \vspace{.1in}
    \caption{Evolution of the parameters of the estimated adjacency matrix. $B^\ast$ denotes the true adjacency matrix and $B^{(n)}$ denotes the parameters of the estimated adjacency matrix after the $n$-th training iteration.}
\end{figure*}

\paragraph{Objective} Given a partially incomplete data set $\mathcal{D}$ sampled from a set of interventional experiments, we would like to learn the structure of the underlying causal $G$, i.e., the directed edge set $E$, by maximizing the likelihood of the observed nodes.

\subsection{The Overall Framework}

Given the data set $\mathcal{D} = \{\x^{(i)}, \r^{(i)}, \u^{(i)}\}_{i=1}^M$, the graph structure is learned by maximizing the sparsity penalized log-likelihood of the non-missing nodes in each sample. That is, 
\begin{equation}
    \max_{\bm{\theta}} \tilde{\mathcal{L}}(\mathcal{D}, \bm{\theta}) = \max_{\bm{\theta}}\sum_{i=1}^M \log p_X(\x_{\Gamma_i}^{(i)}\mid\bm{\theta}) - \lambda R_\text{sp}(\bm{\theta}),
    \label{eq:obj}
\end{equation}
where $R_\text{sp}(\bm{\theta})$ is a regularizer that promotes sparsity in the estimated causal graph. As discussed in section \ref{sec:prob-setup}, it is generally intractable to compute $\log p_X(\x_{\Gamma_i}^{(i)}\mid\bm{\theta})$ and no closed form solution exists. However, \eqref{eq:obj} can be solved using the iterative penalized expectation-maximization (EM) method \citep{chen2014penalized}. Unlike the data imputation algorithms that directly try to sample the missing values, starting with an initial parameter $\bm{\theta}^0$, the EM algorithm alternates between the following two steps in each iteration:

\paragraph{E-step:} In this step, we use the current estimate of the model parameters and the non-missing data to compute the expected log-likelihood of the complete data.
\begin{align}
    Q(\bm{\theta}, \bm{\theta}^t) &= \sum_{i=1}^M \mathbb{E}_{\x_{\Omega_i}^{(i)}\mid \x_{\Gamma_i}^{(i)}; \bm{\theta}^t} \Big[\log p_X(\x_{\Omega_i}^{(i)}, \x_{\Gamma_i}^{(i)}\mid \bm{\theta}) \Big]
     \label{eq:e-step}
\end{align}

\paragraph{M-step:} Here, we maximize the expected log-likelihood of the complete data computed in the previous step with respect to $\bm{\theta}$. That is, 
\begin{equation}
    \bm{\theta}^{t+1} = \arg\max_{\bm{\theta}} Q(\bm{\theta}, \bm{\theta}^t) - \lambda \|\bm{\theta}\|_1.
    \label{eq:m-step}
\end{equation}
We use stochastic gradient-based solvers to solve the above maximization problem. Note that,
\begin{equation}
    \sum_{i=1}^M\log p_X (\x_{\Gamma_i}^{(i)}\mid \bm{\theta}) \geq Q(\bm{\theta}, \bm{\theta}^t) - \text{const}.
    \label{eq:elbo}
\end{equation}
The above equation implies that we maximize a lower bound of the log-likelihood as we update the parameters during the M-step. More details on the derivation of \eqref{eq:elbo} and on the convergence analysis of the MissNODAGS are provided in the supplementary materials (Appendix A).    

\subsection{Computing the Log-Likelihood in E-step}
\label{ssec:likelihood-estep}

Computing the log-likelihood in the E-step can be computationally challenging since from \eqref{eq:px-interv}, $\log p_X(\x_{\Omega_i}^{(i)}, \x_{\Gamma_i}^{(i)}\mid \bm{\theta})$ requires the computation of the following term
$$\log \Big|\det \J_{(\id - \U f)}(\x)\Big|.$$
In the worst case, the Jacobian matrix could require gradient calls in the order of $d^2$. To that end, following \citet{pmlr-v206-sethuraman23a}, we employ contractive residual flows \citep{iresnet, russianroulette} in order to compute the log-determinant of the Jacobian in a tractable manner. 

\paragraph{Modeling the causal function.} We start by restricting the causal function $f$ to be contractive. A function $f$ is said to be contractive if it is a Lipschitz function with Lipschitz constant $L < 1$. It then follows from Banach fixed point \citep{banachfp} that the map $(\id - \U f)$ for interventional experiment $\ec = (\ic, \uc)$ is contractive and invertible. We use neural networks to model the contractive function $f$. As shown by \citet{iresnet}, neural networks can be constrained to be contractive during the training phase by rescaling its layer weights by their corresponding spectral norm \citep{iresnet} after each iteration.

Dependency mask $\bm{M}\sim\{0, 1\}^{d\times d}$ is introduced to explicitly encode the dependencies between the nodes in the graph. Thus the causal function has the following form
$$[f_{\bm{\theta}}(\x)]_i = [\text{NN}_{\bm{\theta}}(\bm{M}_{\ast,i} \odot \x)]_i.$$
The dependency mask is sampled from Gumbel-softmax distribution \citep{jang2016categorical}, $\bm{M} \sim \bm{M}_\phi$ and the parameters $\phi$ are updated during the training (M-step). The sparsity penalty thus becomes $\mathbb{E}_{\bm{M\sim\bm{M}_\phi}}\|M\|_1$.

\paragraph{Computing the log-det of Jacobian.} Using power series expansion, the log-determinant of the Jacobian can be computed using the following unbiased estimator introduced by \citet{iresnet}, 
\begin{align}
    \log \Big|\det \J_{(\id - \U f)}(\x)\Big| &= \log \Big|\det(\I - \U\J_f)(\x)\Big| \nonumber \\
    &= -\sum_{k=1}^\infty \frac{1}{k} \text{Tr}\Big\{\J_{\U f}^k(\x)\Big\}.
    \label{eq:power-series}
\end{align}
The above series is guaranteed to converge since $f$ is contractive. In the current form we still require gradient calls in the order of $d$. We can reduce this further by using the Hutchinson trace estimator \citep{hutchtraceestimator}
\begin{equation}
    \text{Tr}\Big\{\J_{\U f}^k(\x)\Big\} = \mathbb{E}_{\bm{w}} \Big[\bm{w}^\top \J_{\U f}^k(\x)\bm{w}\Big],
    \label{eq:hutch-est}
\end{equation}
where $\bm{w} \sim \mathcal{N}(\bm{0}, \I)$. 

In practice, \eqref{eq:power-series} is computed by truncating the number of terms in the summation to a finite number. This, however, introduces bias in estimating the log-determinant of the Jacobian. In order to circumvent this issue we follow the steps taken by \citet{russianroulette}. The power series expansion is truncated at a random cut-off $n\sim p(N)$, where $p$ is a probability distribution over the natural numbers. In our work, we choose this distribution to be a Poisson distribution, $n\sim\text{Poi}(N)$, where $N$ is treated as a hyperparameter. Each term in the finite power series is then re-weighted to obtain the following estimator for the log-determinant of the Jacobian,
\begin{equation}
    \log\big|\det J_{(\id - \U f)}(\x)\big| = -\mathbb{E}_{n, \bm{w}}\Bigg[\sum_{k=1}^n \frac{\bm{w}^\top J_{\U f}^k(\x)\bm{w}}{k\cdot P(N \geq k)}\Bigg],
    \label{eq:russian}
\end{equation}
where $P$ is the cumulative density function corresponding to $p$. Combining \eqref{eq:px-interv}, \eqref{eq:power-series}, and \eqref{eq:russian}, we finally get
\begin{multline}
     Q(\bm{\theta}, \bm{\theta}^t) \propto \sum_{i=1}^M \mathbb{E}_{\x_{\Omega_i}^{(i)}\mid \x_{\Gamma_i}^{(i)}; \bm{\theta}^t}\bigg\{ p_E\Big([(\id - \U_i f)(\x)]_\uc\Big)\\
        - \mathbb{E}_{n, \bm{w}}\Bigg[\sum_{k=1}^n \frac{\bm{w}^\top J_{\U f}^k(\x)\bm{w}}{k\cdot P(N \geq k)}\Bigg].
        \label{eq:q-final}
\end{multline}

\subsection{MissNODAGS for Linear Gaussian SEM}
\label{ssec:missnodags-linear-sem}

For the case when the structural equations are linear with the noise distribution being Gaussian, i.e., 
\begin{equation}
    \x = \bm{B}^\top \x + \e,
\end{equation}
where $\e \sim \mathcal{N}(0,\bm{\Theta}^{-1})$ and $\bm{B}$ is the weighted adjacency matrix of $G$, we have that $\x$ is also Gaussian distributed with $\x \sim \mathcal{N}(\bm{0}, \bm{\Theta}_X^{-1})$ where
\begin{equation}
    \bm{\Theta}_X = (\I - \bm{B})\bm{\Theta}(\I - \bm{B}^\top),
    \label{eq:prec-x}
\end{equation}
and $\bm{\Theta} = \text{diag}(1/\sigma_1^2, \ldots, 1/\sigma_d^2)$. In order to compute \eqref{eq:q-final} we need to compute the expectation with respect to the posterior distribution of the missing nodes conditioned on the observed nodes, $p_X(\x_{\Omega_i}^{(i)}\mid \x_{\Gamma_i}^{(i)}; \bm{\theta}^t)$. Since $\x$ is a Gaussian random vector, from the properties of Gaussian distribution, the conditional distribution of the missing nodes given the observed nodes also follows a Gaussian distribution, $\x_{\Omega_i}^{(i)}\mid \x_{\Gamma_i}^{(i)} \sim \mathcal{N}(\tilde{\bm{\mu}}^{(i)}, \tilde{\bm{\Theta}}^{(i)})$ such that
\begin{align}
    \tilde{\bm{\Theta}}^{(i)} &= [\bm{\Theta}_X]_{\Omega_i, \Omega_i};\label{eq:prec}\\ 
    \tilde{\bm{\eta}}^{(i)} &= -\tilde{\bm{\Theta}}^{(i)} \x_{\Gamma_i}^{(i)}, \label{eq:eta}
\end{align}
where $\tilde{\bm{\eta}}^{(i)} = \tilde{\bm{\Theta}}^{(i)}\tilde{\bm{\mu}}^{(i)}$.

The conditional expectation in \eqref{eq:q-final} is then estimated by imputing the missing values in each data instance by sampling from the Gaussian distribution parameterized by \eqref{eq:prec} and \eqref{eq:eta}. More details on the imputation procedure can be found in subsection \ref{ssec:impute}. 

Now, we explain how imputation is done when interventions are present. Given an interventional experiment $\ec = (\ic, \uc)$, \eqref{eq:prec-x} can be modified as follows
\begin{equation}
    \bm{\Theta}_{X, \ic} = (\I - \U\bm{B})(\U\bm{\Theta} + (\I - \U))(\I - \U\bm{B}^\top),
    \label{eq:prec-x-inter}
\end{equation}
here we assume that $\x_\ic \sim \mathcal{N}(\bm{0}, \I)$. The rest of the imputation procedure is the same as before with the only difference being that we use $\bm{\Theta}_{X, \ic}$ in place of $\bm{\Theta}_X$ in \eqref{eq:prec}.

Given a set of interventional experiments $\{\ec_k\}_{k=1}^K$, under some additional assumptions, we provide consistency guarantees for MissNODAGS in the form of the following theorem. 

\begin{theorem}
Under the assumptions stated in Appendix A, the global minimizer of \eqref{eq:obj} with a suitable choice of $\lambda$ outputs $\tilde{G} \cong_\ic G$ for linear Gaussian SEM.
\label{thm:consistency}
\end{theorem}

Where $G$ is the true causal graph and $\cong_\ic$ denotes equivalence with respect to quasi-equivalence \citep{ghassami2020characterizing} extended to the interventional setting \citep{pmlr-v206-sethuraman23a} for a family of interventional experiments $\{\ec_k\}_{k=1}^K$. The proof and the underlying assumptions for Theorem \ref{thm:consistency} can be found in Appendix \ref{app:proofs}.

\subsection{MissNODAGS for Nonlinear SEM}

For the general case of \eqref{eq:sem-comb}, it is not easy to compute the posterior distribution of the missing nodes, $p_X(\x_{\Omega_i}^{(i)}\mid \x_{\Gamma_i}^{(i)}; \bm{\theta}^t)$, making the maximization of expected likelihood more difficult. In this case, we propose to approximate the posterior distribution.  

For the E-step, with the assumption that the causal function $f$ is contractive (required for computing the log-likelihood, and the convergence of the SEM when feedback loops are present), first-order Taylor series approximation is used to approximate the nonlinear SEM. Hence we have,
\begin{equation}
    \x \approx \x_0 + \J_f(\x_0)(\x - \x_0) + \e.
    \label{eq:sem-lin-approx}
\end{equation}
Here we choose $\x_0 = \bm{0}$. Since $f$ is contractive, that is, $\norm{f(\x) - f(\bm{y}}) \leq \norm{\x - \bm{y}}$ for $\x, \bm{y} \in \mathbb{R}^d$, we have that 
\begin{align*}
    \norm{\J_f(\bm{0})\cdot\x - f(\x)} \leq \norm{\J_f(\bm{0})\cdot\x} + \norm{f(\x)}
\end{align*}
From the contractivity of $f$, it follows that $\norm{\J_f(\bm{z})} \leq 1$ for all $\bm{z} \in \mathbb{R}^d$, and $\norm{f(\x) - f(\bm{0})} = \norm{f(\x)} \leq \norm{\x}$. Therefore, 
$$\norm{\J_f(\bm{0})\cdot\x - f(\x)} \leq 2\norm{\x}.$$
Hence, the approximation error increases as the point $\x$ moves away from the zero vector. Since the mean of $\x$ is zero, the first-order Taylor series approximation serves as a good proxy for the causal function $f$ for the purpose of imputing the missing values. 

With the latent distribution set to be Gaussian, $\e \sim \mathcal{N}(\bm{0}, \bm{\Theta}^{-1})$, \eqref{eq:sem-lin-approx} implies that $\x$ is Gaussian distributed. Treating $\J_f^\top$ as the new weighted adjacency matrix, then, $\x \sim \mathcal{N}(\bm{0}, \bm{\Theta}_X^{-1})$ where 
\begin{equation}
    \bm{\Theta}_X = (\I - \bm{J}_f^\top)\bm{\Theta}(\I - \bm{J}_f).
\end{equation}
This again implies that the conditional distribution is Gaussian and hence we follow the same procedure discussed in subsection \ref{ssec:missnodags-linear-sem} for imputing the missing nodes in the E-step. Note that the linear approximation is only used for sampling from the conditional distribution of the missing nodes given the non-missing nodes and not in computing the likelihood function.

At this point, it is important to note that, while our overall framework is inspired by that of \cite{gao2022missdag}, the imputation technique used in each iteration of MissNODAGS is significantly different than the one in \cite{gao2022missdag}, as seen in subsection \ref{ssec:impute}.

\subsection{Details on Missing value Imputation}
\label{ssec:impute}

\begin{algorithm}[t] 
\caption{\textsc{Impute-Gaussian}}
\label{alg:data-imputation}
\begin{algorithmic}[1]
\Require{Minibatch data $\mathcal{B} = \{\x_i, \r_i, \uc\}_{i=1}^{N_B}$, current estimate of causal graph $\bm{B}^{(t)}$.} 
\Ensure{Imputed data $\tilde{\mathcal{B}} = \{\tilde{\x}_i, \r_i, \uc\}_{i=1}^{N_B}$}.
\vspace{0.2cm}
\State { Compute $\bm{\Theta}_X = (\I - \U\bm{B})(\U\bm{\Theta} + (\I - \U))(\I - \U\bm{B}^\top).$ } 
    
    \For{$i = 1$ to $N_B$} 
        \State {Construct $\bm{P}_i$ such that $\bm{P}_i\x^{(i)} = [\x_\Omega^{(i)}, \x_\Gamma^{(i)}]$.} 
        
        \State {Compute \textbf{Cholesky decomposition}: $$\bm{M}^\top\M = \bm{P}_i\bt\bm{P}_i^\top$$}
        \State {Solve for $\tilde{\x}_\Omega^{(i)}$, such that $\bm{M}_{\Omega, \Omega}\tilde{\x}_\Omega^{(i)} = \bm{z} - \M_{\Omega,\Gamma}\x_\Gamma^{(i)}$, where $\bm{z} \sim \mathcal{N}(\bm{0}, \I)$.}
        \State {Set $\tilde{\x}^{(i)} = [\x_\Gamma^{(i)}, \tilde{\x}_\Omega^{(i)}]$}. 
    \EndFor
    
    \Return {$\tilde{\mathcal{B}} = \{\tilde{\x}_i, \r_i, \uc\}_{i=1}^{N_B}$.}
\end{algorithmic}
\end{algorithm}
\begin{figure*}[t]
    \centering
    \includegraphics[width=0.7\linewidth]{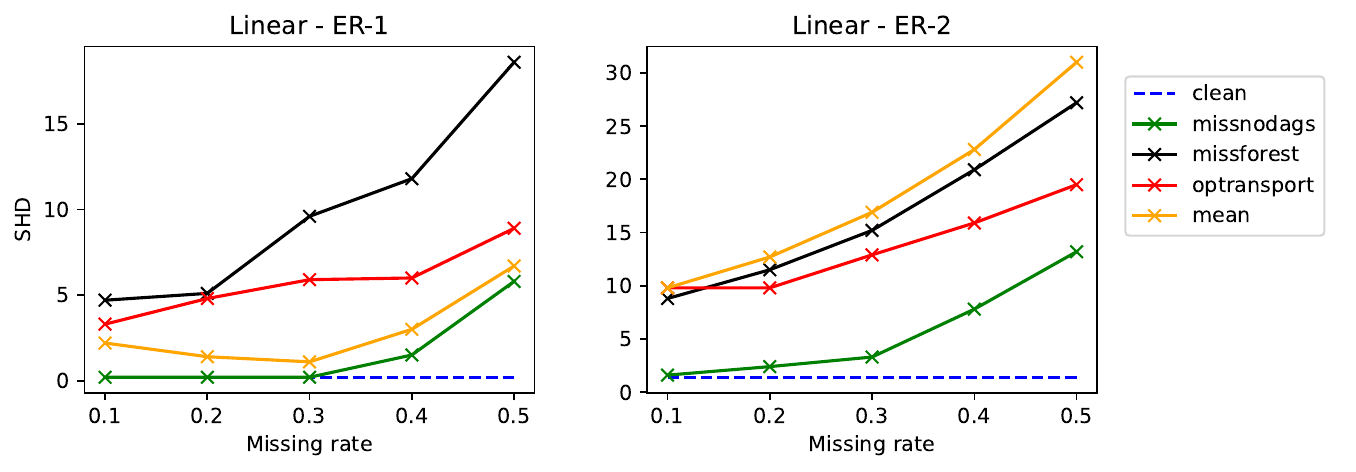}
    \vspace{.1in}
    \caption{Results on recovery of linear Gaussian SEM with $d=20$ nodes.}
    \label{fig:linear-results}
\end{figure*}
We now detail how the missing values are imputed given a batch of data $\mathcal{B} = \{\x^{(i)}, \bm{r}^{(i)}, \uc\}_{i=1}^{N_B}$, where $N_B$ is the batch size. Here, we assume that all the samples in a batch belong to the same interventional experiment. Consider a sample $\x \in \mathcal{B}$, one way to sample the missing values could be as follows, 
\begin{equation}
    \x_{\Omega} = \tilde{\bm{\mu}} + \tilde{\bm{\Theta}}^{-1/2}\bm{z}, \quad \bm{z} \sim \mathcal{N}(\bm{0}, \I).
    \label{eq:miss-sample}
\end{equation}
While $\tilde{\bm{\Theta}}$ can be computed readily from \eqref{eq:prec} and (14) in the main text, $\tilde{\bm{\Theta}}^{-1}$ can be expensive to compute. Moreover, $\{\Omega_i, \Gamma_i\}$ can vary from one sample to another. Alternatively, we can re-write \eqref{eq:miss-sample} by left multiplying both sides of the equation with $\tt$, 
\begin{equation}
    \tt^{1/2}\x_\Omega = \tt^{1/2}\tilde{\bm{\mu}} + z. 
\end{equation}
This circumvents the need for computing the inverse of $\tt$ by posing the sampling of $\x_\Omega$ as a solution to a triangular system of linear equations, which is less computationally expensive. Furthermore, using Cholesky decomposition, 
\begin{equation}
\bt_X = \bm{M}^\top \bm{M},
\end{equation}

where $\bm{M}$ is an upper triangular matrix. Since $\bt$ is a positive definite matrix we then have  $\bt^{1/2} = \bm{M}$. Assuming the $\x = [\x_\Omega, \x_\Gamma]$, i.e., the missing nodes are in the top half, we then have $\tt^{1/2} = \bm{M}_{\Omega, \Omega}$ and $\tt^{1/2}\bm{\tilde{\mu}} = -\bm{M}_{\Omega,\Gamma} \x_\Gamma$. To see why this is true, note that 
\begin{align*}
\bt_X = &
\begin{bmatrix}
    \M_{\Omega, \Omega}^\top & 0\\
    \M_{\Omega, \Gamma}^\top & \M_{\Gamma, \Gamma}^\top
\end{bmatrix}
\begin{bmatrix}
    \M_{\Omega, \Omega} & \M_{\Omega, \Gamma}\\
     0 & \M_{\Gamma, \Gamma}
\end{bmatrix}\\
= &
\begin{bmatrix}
    \M_{\Omega, \Omega}^\top \M_{\Omega, \Omega} & \M_{\Omega, \Omega}^\top \M_{\Omega, \Gamma}\\
    \M_{\Omega, \Gamma}^\top \M_{\Omega, \Omega} & \M_{\Omega, \Gamma}^\top \M_{\Omega, \Gamma} + M_{\Gamma, \Gamma}^\top M_{\Gamma, \Gamma}     
\end{bmatrix}
\end{align*}

Finally, $\x_\Omega$ can be sampled by solving the following triangular system of equations
\begin{equation}
    \bm{M}_{\Omega, \Omega}\x_\Omega = \bm{z} - \M_{\Omega,\Gamma}\x_\Gamma, \quad \bm{z} \sim \mathcal{N}(\bm{0}, \I).
\end{equation}
When the missing nodes are not in the top half of $\x$ then, we permute the nodes such that $\x = [\x_\Omega, \x_\Gamma]$. Algorithm \ref{alg:data-imputation} shows the overall sampling procedure. 

For the case when the SEM is nonlinear, the nonlinear function is approximated using the first-order Tayler series approximation. The Jacobian of the function $\J_{\U f}$ is then treated as the weighted adjacency matrix. 

\section{EXPERIMENTS}
\label{sec:experiments}

We tested MissNODAGS on both synthetic and real-world data sets. As baselines, we first impute the missing values in the data set with state-of-the-art imputation methods such as Optimal Transport (OP) imputation \citep{muzellec2020missing} (\texttt{optransport}), MissForest imputation \citep{stekhoven2012missforest} (\texttt{missforest}), and mean imputation (\texttt{mean}) followed by learning the causal graph from the imputed data using NODAGS-Flow \citep{pmlr-v206-sethuraman23a}. For mean imputation, we replace the missing value with its mean computed from the samples where the node is observed.

In all our experiments, we set the missingness mechanism to be independent of the node observations. The probability that a node value is missed (missing rate) is varied between 0.1 and 0.5 in steps of 0.1. On the synthetic data, the cyclic-directed graphs with $d = 20$ nodes were generated randomly using the Erd\H{o}s-R\'enyi random graph model, where the expected edge density was varied between 1 and 2 (denoted as ER-1, and ER-2 respectively). Once the graphs are generated, the self-loops (if present) are removed. 

\subsection{Synthetic Data}
\begin{figure*}[t]
    \centering
    \includegraphics[width=0.7\linewidth]{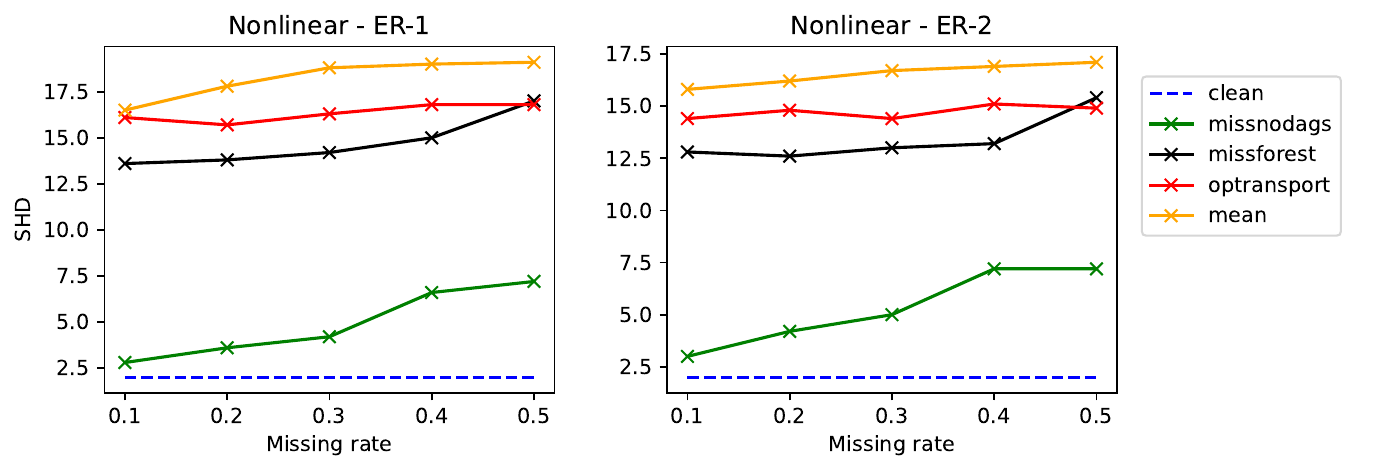}
    \vspace{.1in}
    \caption{Results on recovery of nonlinear Gaussian SEM with $d=20$ nodes.}
    \label{fig:nonlinear-results}
\end{figure*}

\subsubsection{Linear Gaussian SEM} 

The edge weights were sampled from a uniform distribution with the span $(-0.6, -0.25)\cup(0.25, 0.6)$. The weighted adjacency matrix was then rescaled to ensure that the weighted adjacency matrix was contractive. The data consists of single-node interventions over all the nodes in the graph. For each intervention, 100 observations were sampled. Each intervened node was sampled from a zero mean and unit variance normal distribution. The latent noise distribution $p_E(\e)$ was set to be Gaussian, $\e \sim \mathcal{N}(\bm{0}, \sigma^2\I)$, where $\sigma = 0.25$.

The performance of MissNODAGS and the baselines were evaluated using the \emph{Structural Hamming Distance} (SHD) as an error metric. SHD measures the total operations (deletion, addition, and reversal) on the edges required to convert the estimated graph structure to the true graph structure. Both MissNODAGS and NODAGS-Flow were trained for 100 epochs on the generated data. We repeat the experiments over 10 random graphs and the mean SHD across the trials are reported in Figure \ref{fig:linear-results}. The blue dashed line in the figure corresponds to NODAGS-Flow trained on the complete data (\texttt{clean}).

When the edge density is 1, from Figure \ref{fig:linear-results} we can see that MissNODAGS outperforms all of the chosen baselines, NODAGS-Flow with mean imputation being a close second. In fact, when the missing rate is less than 0.4, MissNODAGS matches the performance of NODAGS-Flow on the complete data. 

Similarly, when the edge density is 2, the overall performance of all the methods is reduced slightly. However, MissNODAGS still outperforms all the baselines with the difference in performance being more significant. This experimentally showcases the benefit of alternating between data imputation and graph learning in the cyclic linear Gaussian setting.

\subsubsection{Nonlinear SEM. }

For the nonlinear SEM, we consider the causal function $f$ to be of the following form, 
\begin{equation}
f(\x) = \tanh(\bm{W}^\top \x).
\label{eq:exp-func}
\end{equation}
The non-zero entries of $\bm{W} \in \mathbb{R}^{d\times d}$ are sampled from a uniform distribution with post-scaling to ensure that $f$ is contractive. Here, the Lipschitz constant of $f$ is set to 0.9. As in the previous case, the data set consists of single-node interventions over all the nodes in the graph with each intervention containing 100 samples. In order to generate the data, the noise variables are again sampled from Gaussian distribution. 

Neural networks with a single hidden layer and tanh activation functions are used to learn the causal function $f$. The objective function \eqref{eq:m-step} is maximized using Adam optimizer \citep{kingma2014adam}. Both MissNODAGS and NODAGS-Flow are trained for 100 epochs and the mean SHD over 10 trails are reported in Figure \ref{fig:nonlinear-results}.

As seen in Figure \ref{fig:nonlinear-results}, we observe a similar trend in the performance of MissNODAGS and the baselines for both the values of edge density. That is, MissNODAGS outperforms all the baselines, with the difference in performance being more significant than in the linear case. Interestingly, there isn't a notable drop in performance as the edge density increases. On the other hand, the baselines have comparable performances in this setting with MissForest performing the best amongst the baselines.

\begin{figure}[H]
    \centering
    \includegraphics[width=0.6\linewidth]{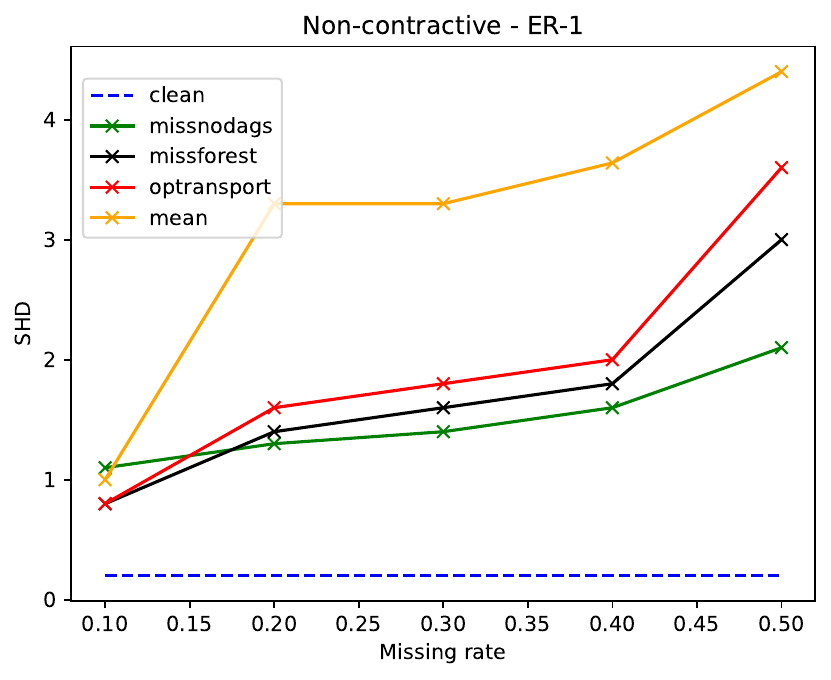}
    \caption{Results on non-contractive data}
    \label{fig:non-contractive}
\end{figure}

\begin{figure*}[t]
    \centering
    \includegraphics[width=\linewidth]{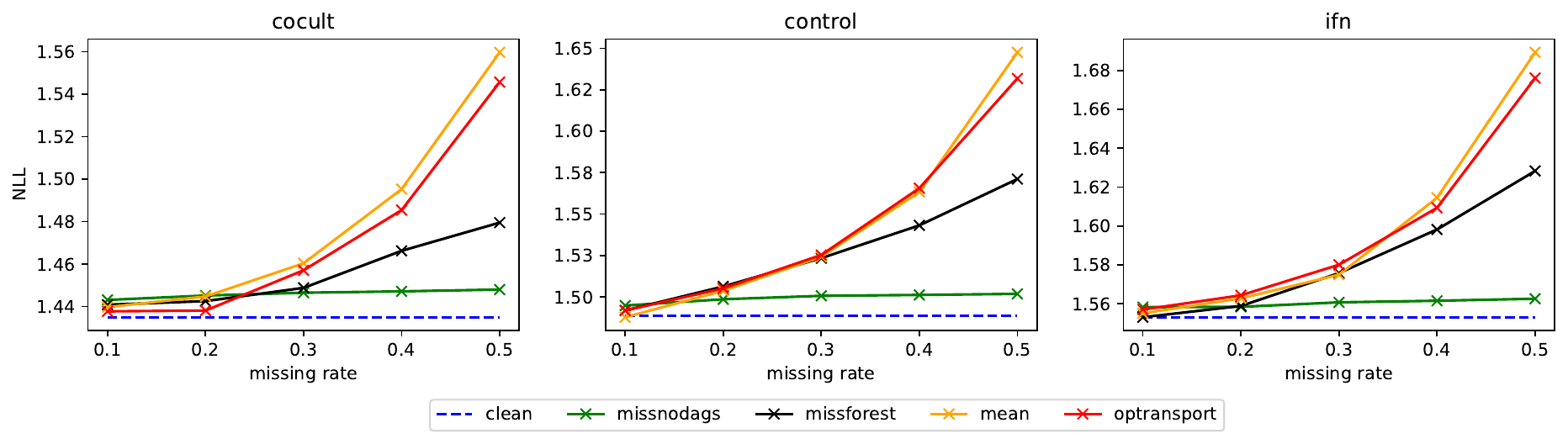}
    \caption{Results on Perturb-CITE-seq data set \citep{frangieh2021multimodal}.}
    \label{fig:perturb-cite-seq-results}
\end{figure*}

Additionally, we compare the performance of MissNODAGS with the chosen baselines on non-contractive non-linear data. To that end, the weights matrix in \eqref{eq:exp-func} is scaled to be non-contractive, the Lipschitz constant of $f(\x)$ was set to 2.0. In this case, the performance of MissNODAGS is much more closer to the baseline than in the contractive setting, Figure \ref{fig:non-contractive}. We attribute this to the approximation error while imputing the missing values in each iteration, as the linear approximation becomes poor as the function becomes increasingly non-linear.

\begin{table}[H]
    \centering
    \caption{Comparison between MissNODAGS and \cite{gao2022missdag}.}
    \begin{tabular}{c|c|c}
    \hline\hline
         Method & SHD & Time (sec)\\
         \hline\hline
         MissNODAGS & \textbf{0.4} & \textbf{1.7}\\
         mMissDAG (\cite{gao2022missdag}) &  1.5 & 77.1\\
         \hline
    \end{tabular}
    \label{tab:mcmc}
\end{table}

Finally, we also evaluated the performance of MissNODAGS with the method proposed by \cite{gao2022missdag} (MissDAG), which is capable of handling incomplete non-linear data. MissDAG uses an MCMC sampling based approach (rejection sampling) for imputing the missing values in every iteration of the algorithm. This makes the overall algorithm quite slow. Hence, we restrict the comparison to graphs with 3 nodes, with expected edge density equal to 1. Since MissDAG isn't compatible with cyclic causal graphs, we modify the baseline to allow for interventions and feedback loops (denoted as mMissDAG). In this case, we set the causal mechanism to be contractive and fix the missing rate to 0.2. 

As seen from Table \ref{tab:mcmc}, MissNODAGS outperforms the MCMC based approach proposed by \cite{gao2022missdag}, while also being considerably faster. For compute time comparison, we report the time taken per training epoch to train each model for 50 epochs on a Nvidia GTX6000 GPU.

\subsection{Real World Transcriptomics Data}

Here we present an experiment focused on learning causal graph structure corresponding to a gene regulator network from a gene expression data with genetic interventions. In particular, we focus on the Perturb-CITE-seq dataset \citep{frangieh2021multimodal}, a type of data set that allows one to study causal relations in gene networks at an unprecedented scale. It contains gene expressions taken from 218,331 melanoma cells split into three cell conditions: (i) control (57,627 cells), (ii) co-culture (73,114 cells), and (iii) interferon (INF)-$\gamma$ (87,590 cells).

Due to the practical and computational constraints, of the approximately 20,000 genes in the genome, we restrict ourselves to a subset of 61 genes, following the experimental setup of \citet{pmlr-v206-sethuraman23a}, summarized in Table \ref{tab:genes}. We then took all the single-node interventions corresponding to the 61 genes. Each cell condition is treated as a separate data set over which the models were trained separately. 

Since the data set does not provide a ground truth causal graph, it is not possible to directly compare the performance using SHD. Instead, we compare the performance of the causal discovery methods based on its predictive performance over unseen interventions. To that end, we perform a 90-10 split on the three data sets. The smaller set is treated as the test set, which is then used for performance comparison between MissNODAGS and the baselines. As a performance metric, we use the predicted negative log-likelihood (NLL) over the test set after training the models for 100 epochs. The results are reported in Figure \ref{fig:perturb-cite-seq-results}.

When the missing rate is low ($< 0.2$), all the methods exhibit comparable performance and match that of NODAGS-Flow trained on complete data. However, as the missing rate increases the relative performance starts to diverge and MissNODAGS outperforms all the baseline methods. In fact, the NLL on the test set remains at a relatively steady level even when the missing rate increases and is close to clean NODAGS-Flow. We also observe the same behavior across the three cell conditions.

\begin{table*}[t]
    \centering
    \caption{The 61 genes chosen from Perturb-CITE-seq dataset \citep{frangieh2021multimodal}.}
    \vspace{0.2cm}
    \begin{tabular}{|llllllllll|}
    \hline
        ACSL3 & ACTA2 & B2M & CCND1 & CD274 & CD58 & CD59 & CDK4 & CDK6  & ~ \\
        CDKN1A & CKS1B & CST3 & CTPS1 & DNMT1 & EIF3K & EVA1A & FKBP4 & FOS  & ~ \\ 
        GSEC & GSN & HASPIN & HLA-A & HLA-B & HLA-C & HLA-E & IFNGR1 & IFNGR2  & ~ \\ 
        ILF2 & IRF3 & JAK1 & JAK2 & LAMP2 & LGALS3 & MRPL47 & MYC & P2RX4  & ~ \\ 
        PABPC1 & PAICS & PET100 & PTMA & PUF60 & RNASEH2A & RRS1 & SAT1 & SEC11C  & ~ \\ 
        SINHCAF & SMAD4 & SOX4 & SP100 & SSR2 & STAT1 & STOM & TGFB1 & TIMP2  & ~ \\ 
        TM4SF1 & TMED10 & TMEM173 & TOP1MT & TPRKB & TXNDC17 & VDAC2 & ~ &   & ~ \\ \hline
    \end{tabular}
    \label{tab:genes}
\end{table*}

\section{DISCUSSION}

In this work, we proposed MissNODAGS, a novel framework based on expectation maximization (EM) that is capable of learning nonlinear cyclic causal graphs from incomplete interventional data. We analyzed two cases for the structural equations models: (i) Linear Gaussian SEM, and (ii) Nonlinear SEM, and presented how our framework handles the two scenarios. In particular, for the nonlinear SEM setting, we showed that approximating the SEM via first-order Taylor series approximation for imputing the missing values achieves promising results in recovering the causal graph structure. Experiments on synthetic data also validate the efficacy of our approach in learning nonlinear cyclic causal graphs from incomplete data. 

We also presented an experiment on a real-world gene expression data (Perturb-CITE-seq), where we used the learned causal graph to predict the likelihood of unseen interventions. Additionally, we provide consistency guarantees for our framework for the linear Gaussian SEM setting. 

Interesting directions for future research include: (1) incorporating a realistic measurement noise model into the structural equations to improve the performance on real-world data sets, (2) scaling the current framework to larger graphs using low-rank models and variation inference based techniques, and (3) allowing for unobserved confounders within the modeling assumptions. 

\bibliography{references}

\pagebreak

\appendix
\onecolumn

The appendix is organized as follows: in Section \ref{app:proofs} we analyze the convergence of MissNODAGS and provide the proof of Theorem 1,  followed by implementation details of MissNODAGS and the baselines in Section \ref{app:imp-details}.

\section{PROOFS}
\label{app:proofs}

\subsection{Convergence analysis of MissNODAGS}
\label{app:convegence}

Here we provide the convergence analysis of MissNODAGS. Our analysis relies on the convergence analysis of the EM algorithm \citep{wu_em_convergence, Friedman1998TheBS}. The crux of the analysis depends on establishing that the total log-likelihood of the non-missing nodes in the data set either increases or stays the same in each iteration of the algorithm. That is, 
\begin{equation}
\sum_{i=1}^M \log p_X(\x_{\Gamma_i}^{(i)} \mid \bm{\theta}^{t+1}) \geq \sum_{i=1}^M \log p_X(\x_{\Gamma_i}^{(i)} \mid \bm{\theta}^{t}).
\label{eq:requirement}
\end{equation}
To that end, note that 
\begin{align*}
\sum_{i=1}^M\log p_X(\x_{\Gamma_i}^{(i)} \mid \bm{\theta}) &= \sum_{i=1}^M\log p_X(\x_{\Gamma_i}^{(i)}\x_{\Omega_i}^{(i)} \mid \bm{\theta}) \\
&\qquad \qquad- \sum_{i=1}^M\log p_X(\x_{\Omega_i}^{(i)} \mid \x_{\Gamma_i}^{(i)}, \bm{\theta})\\
&= \underbrace{\sum_{i=1}^M\mathbb{E}_{\x_{\Omega_i}^{(i)}\mid \x_{\Gamma_i}^{(i)}; \bm{\theta}^t}\log p_X(\x_{\Gamma_i}^{(i)}\x_{\Omega_i}^{(i)} \mid \bm{\theta})}_{ = Q(\bm{\theta}, \bm{\theta}^t)}\\
&\qquad - \sum_{i=1}^M\mathbb{E}_{\x_{\Omega_i}^{(i)}\mid \x_{\Gamma_i}^{(i)}; \bm{\theta}^t}\log p_X(\x_{\Omega_i}^{(i)} \mid \x_{\Gamma_i}^{(i)}, \bm{\theta}).
\end{align*}
The first term on the RHS in the above equation is nothing but $Q(\bm{\theta}, \bm{\theta}^t)$. This is maximized in the M-step, that is, $Q(\bm{\theta}, \bm{\theta}^{t+1}) \geq Q(\bm{\theta}, \bm{\theta}^t)$. On the other hand, 
\begin{equation*}
\sum_{i=1}^M\mathbb{E}_{\x_{\Omega_i}^{(i)}\mid \x_{\Gamma_i}^{(i)}; \bm{\theta}^t}\log \frac{p_X(\x_{\Omega_i}^{(i)} \mid \x_{\Gamma_i}^{(i)}, \bm{\theta}^{t+1})}{p_X(\x_{\Omega_i}^{(i)} \mid \x_{\Gamma_i}^{(i)}, \bm{\theta}^t)} = -D_{KL}\Big(p_X(\x_{\Omega_i}^{(i)} \mid \x_{\Gamma_i}^{(i)}, \bm{\theta}^t)\big\|p_X(\x_{\Omega_i}^{(i)} \mid \x_{\Gamma_i}^{(i)}, \bm{\theta}^{t+1})\Big) \leq 0.
\end{equation*}
Therefore, at the end of the M-step \eqref{eq:requirement} is satisfied. Similar to the previous results on EM convergence \citep{wu_em_convergence, Friedman1998TheBS}, MissNODAGS reaches a stationary point of the optimization objective.

\subsection{Consistency of MissNODAGS}

In this subsection, we provide proof of consistency of the MissNODAGS estimator for the linear Gaussian SEM setting. That is, 
\begin{equation}
\x = \B^\top\x + \e,
\label{eq:lin-sem}
\end{equation}
where $\e\sim\mathcal{N}(\bm{0}, \bm{\Omega})$, and $\bm{\Omega}$ is the covariance matrix of $\e$ which we assume to be diagonal. Under some conditions, we show that in the population setting, MissNODAGS estimator returns a graph that is equivalent to the true causal graph under the notion of interventional quasi-equivalence \citep{pmlr-v206-sethuraman23a}, which is an extension of quasi-equivalence \citep{ghassami2020characterizing} to the interventional setting. 

In this case, $\x$ is Gaussian distributed with its precision matrix given by
$$\bm{\Theta} = (\I - \bm{B})\bm{\Theta}(\I - \bm{B}^\top).$$
Given a family of interventional experiments $\{\mathcal{I}_k\}_k$, under the interventions $J \in \{\mathcal{I}_k\}_k$, the precision matrix then becomes 
$$\bm{\Theta}_J = (\I - \U\bm{B})(\U\bm{\Theta} + (\I - \U))(\I - \U\bm{B}^\top).$$

Let $\Theta(G)$ (similarly $\Theta_J(\B)$) denote the set of distributions generated by the graph $G$ over all the possible choices of $\B$ and $\bm{\Omega}$ (for an interventional setting $J$). Let $\eta^{\delta_\text{max}}$ denote the Hausdorff measure of all such parameterized distributions, where $\delta_\text{max}$ is the associated Hausdorff dimension. As defined by \citet{pmlr-v206-sethuraman23a}, a distributional constraint is a constraint that is enforced by the structure of $G$ on the generated distribution. Furthermore, a hard constraint is a distributional constraint for which the set of values satisfying the constraint has zero measure with respect to $\eta^{\delta_\text{max}}$. The set of all hard constraints enforced by $G$ is denoted by $H(G)$. For the sake of completeness, we provide the following definition taken from \citet{pmlr-v206-sethuraman23a} and \citet{ghassami2020characterizing} which would be used towards proving Theorem 1. 

\begin{define}[Interventional quasi-equivalence \citep{pmlr-v206-sethuraman23a}]
    Let $G_1, G_2$ be two directed graphs and $\{\mathcal{I}_k\}_k$ a family of interventional experiments. Let $\delta_J$ denote the Hausdorff dimension of $\Theta(G_1^J) \cup \Theta(G_2^J)$ for every $J \in \{\mathcal{I}_k\}_k$ and denote by $\eta^{\delta_J}$ the associated Hausdorff measure. We define $G_1$ and $G_2$ as \emph{interventionally quasi-equivalent}, denoted by $G_1\cong_{\mathcal{I}} G_2$, if $\eta^{\delta_J}\big(\Theta(G_1^J)\cap\Theta(G_2^J)\big)\neq 0$ for all $J \in \{\mathcal{I}_k\}_k$.
\label{define:inter-quasi-equiv}
\end{define}

Where $G^J$ denotes the mutilated graph corresponding to the interventional setting $J$. An implication of Definition \ref{define:inter-quasi-equiv} is that if $G_1$ and $G_2$ are interventionally quasi-equivalent, for a family of interventions $\{\ic_k\}_k$, then they share the same hard constraints for all $J \in \{\ic_k\}_k$. 

\begin{define}[Generalized Faithfulness \citep{ghassami2020characterizing}]
    A distribution $\Theta$ is said to be generalized faithful (g-faithful) to the graph $G$ if $\Theta$ satisfies a hard constraint $\kappa$ if and only if $\kappa \in H(G)$. 
\label{define:g-faithful}
\end{define}

Following \citet{pmlr-v206-sethuraman23a}, we make the following set of assumptions. 

\begin{assume}
    For a family of interventional experiments $\{\ic_k\}_k$, the interventional distributions $\Theta_J$ are g-faithful to the corresponding ground truth interventional graph $G^J$, for all $J \in \{\ic_k\}_k$.
    \label{assume:g-faithful}
\end{assume}
\begin{assume}
    If there exists directed graphs $G$ and $G_1$ such that $H(G_1) \subseteq H(G)$, and $|E(G_1)| \leq |E(G)|$, then $H(G_1) = H(G)$. 
    \label{assume:sparsity}
\end{assume}
We then consider our EM objective
\begin{equation}
    \max_{\bm{\phi}, \bm{\psi}}\mathbb{E}_{\B\sim\B_{\bm{\phi}}}\mathbb{E}_{\bo\sim\bo_{\bm{\psi}}} \Big[\mathcal{L}(\B, \mathcal{D}, \bo) - \lambda R_\text{sp}(\B)\Big],
    \label{eq:objective}
\end{equation}
where $\mathcal{L}(\B, \mathcal{D}, \bo)$ denotes the likelihood of the non-missing nodes in the dataset $\mathcal{D} = \{\x^{(i)}, \r^{(i)}, \uc^{(i)}\}_{i=1}^M$. $\B_{\bm{\phi}}$ denotes a distribution over the space of $\B$ parameterized by $\bm{\phi}$, similarly, $\bo_{\bm{\psi}}$ denotes a distribution over the space of $\bo$ parameterized by $\bm{\psi}$. 
\begin{assume}
    Given a family of interventional experiments $\{\ic_k\}_k$, the distribution over the space of graphs $\B_{\bm{\phi}}$ contains at least one representative of $\B_\ast$, the corresponding graph $G_\ast$ such that $\bt_J \in \Theta(G_\ast^J)$ for all $J \in \{\ic_k\}_k$.
    \label{assume:well-specified}
\end{assume}

\citet{pmlr-v206-sethuraman23a} consider the case where the data set is complete, this allows for direct maximization of the likelihood function, and hence, the consistency of the maximum likelihood estimator along with the stated assumptions ensures that their approach was consistent. However, in the presence of missing data, direct maximization of the likelihood of the non-missing nodes is not straightforward. As seen in subsection \ref{app:convegence}, the EM algorithm we employ, in general, can only attain a stationary point of the objective function. While this is true for the general case, it is possible that under certain regularity conditions, and when missing rate $\rho$ is bounded above, when the initial parameters $\bm{\theta}^0 = \{\bm{\phi}^0, \bm{\psi}^0\}$ are suitably chosen, the EM algorithm converges to global maxima in the population setting \citep{stat_guarantee_em}, see corollary 3 in \citet{stat_guarantee_em}. 

\begin{assume}
    The initial parameters $\bm{\theta}^0 = \{\bm{\phi}^0, \bm{\psi}^0\}$ satisfy the requirements of Corollary 3 in \citet{stat_guarantee_em}, and $\rho$ is bounded above. 
    \label{assume:good-start}
\end{assume}

\subsubsection{Proof of Theorem 1}

Under the assumptions \ref{assume:g-faithful}, \ref{assume:sparsity}, \ref{assume:well-specified}, and \ref{assume:good-start}, we now show that the global minimizer of \eqref{eq:objective} with a suitable choice of $\lambda$ outputs $\tilde{G} \cong_\ic G$. 

\begin{proof}
Let $G$ denote the ground truth graph and $\bt_J$ the interventional distributions corresponding to the interventional setting $J\in\{\mathcal{I}_k\}_k$ associated with $G$. Let $\B$ and $\bo$ denote the edge weights of $G$ and the intrinsic noise variances respectively. 

We choose the penalty coefficient $\lambda$ such that the likelihood term dominates asymptotically, from assumption \ref{assume:good-start}, the mass of the distributions $\B_{\phi}$ and $\bo_\psi$ will eventually concentrate around the maximum likelihood estimator. If not, the \eqref{eq:objective} would still be sub-optimal. Pick any such pair of parameters $(\hat{\B}, \hat{\bo})$. From assumption \ref{assume:good-start} and optimiality of maximum likelihood estimator, we have that $(\I - \U_J\hat{\B})\hat{\bo}_J^{-1}(\I - \U_J\hat{\B})^{\top} = \bt_J$ for all $J \in \{\mathcal{I}_k\}_k$. Let us denote the directed graph corresponding to $\hat{\B}$ as $\hat{G}$. Since $\bt_J \in \Theta(\hat{G}^J)$ for all $J \in \{\mathcal{I}_k\}_k$, $\bt_J$ must satisfy all the distributional constraints of $G^J$ for all $J \in \{\mathcal{I}_k\}_k$. Therefore from Assumption \ref{assume:g-faithful}, we have that $H(\hat{G}^J) \subseteq H(G^J)$ for all $J \in \{\mathcal{I}_k\}_k$.

Since $R_\text{sp}$ in the objective promotes sparse solutions, we have $|E(\hat{G})| \leq |E(G)|$, otherwise, $G$ would be the solution to the objective. Hence, from assumption \ref{assume:sparsity} we have that $H(\hat{G}^J) = H(G^J)$, for all $J \in \{\mathcal{I}_k\}_k$. Therefore  $\hat{G} \cong_{\mathcal{I}} G$.
\end{proof}

\section{IMPLEMENTATION DETAILS}
\label{app:imp-details}

In this section, we present the implementation details of MissNODAG, the baselines, and details of the experimental setup.

\subsection{MissNODAGS}

We implemented our framework using the Pytorch library in Python and the code used in running the experiments can be found in the \texttt{codes} folder within the supplementary materials. We plan to make the code publicly available on GitHub upon publication of the paper. 

Starting with an initialization of the model parameters $\bm{\theta}^0$, we alternate between the E-step and M-step until the parameters converge. In the E-step, Algorithm \ref{alg:data-imputation} is used for imputing the missing data, followed by maximizing the expected likelihood of the non-missing nodes in the M-step. We follow the same setup as \citet{pmlr-v206-sethuraman23a} for modeling the causal functions, i.e., neural networks (NN) along with dependency mask with entries parameterized by Gumbel-softmax distribution, and for computing the log-determinant of the Jacobian, i.e., power series expansion followed by Hutchinson trace estimator. The final objective in the M-step is maximized using Adam optimizer \citep{kingma2014adam}.

The learning rate in all our experiments was set to $10^{-2}$. The neural network models used in our experiments contained one multi-layer perceptron layer. No nonlinearities were added to the neural networks for the linear SEM experiments. We used tanh activation for the nonlinear SEM experiments and ReLU activation for the experiments on the perturb-CITE-seq data set. The regularization constant $\lambda$ was set to $10^{-2}$ for the synthetic experiments and $10^{-3}$ for the perturb-CITE-seq experiments. All experiments were performed on NVIDIA RTX6000 GPUs.

\subsection{Baselines}

For the baseline NODAGS-Flow, we modify the code base provided by \citet{pmlr-v206-sethuraman23a} to use the imputed samples for maximizing the likelihood. The hyperparameters of NODAGS-Flow was set to the values described in the previous subsection. 

For mean imputation, in each sample the missing nodes are imputed using its mean computed from the sample where the corresponding node was observed in a batch of data. The implementation is available within the \texttt{code} folder in the supplementary materials. MissForest imputation is performed using the publicly available python library \texttt{missingpy}. We use the codebase provided by \citet{muzellec2020missing} for optimal transport imputation, which we include within the \texttt{code} folder. The default parameters are used for both Missforest and optimal transport imputation. 
\end{document}